\documentclass[10pt,twocolumn,letterpaper]{article}

\usepackage[pagenumbers]{cvpr} 

\usepackage{graphicx}
\usepackage{amsmath}
\usepackage{amssymb}
\usepackage{pifont}
\usepackage{booktabs}

\usepackage{wrapfig}
\usepackage{array}
\usepackage{footnote}
\usepackage[ruled,boxed,lined]{algorithm2e}

\newcommand{\cmark}{\ding{51}}%
\newcommand{\xmark}{\ding{55}}%

\usepackage[pagebackref,breaklinks,colorlinks]{hyperref}

\usepackage[capitalize]{cleveref}
\crefname{section}{Sec.}{Secs.}
\Crefname{section}{Section}{Sections}
\Crefname{table}{Table}{Tables}
\crefname{table}{Tab.}{Tabs.}

\def\cvprPaperID{*****} 
\def\confName{CVPR}
\def\confYear{2022}

\begin{document}

\title{IGLU Gridworld: Simple and Fast Environment for Embodied Dialog Agents}

\author{Artem Zholus\thanks{Corresponding author: \texttt{zholus.aa@phystech.edu}}  \thanks{Moscow Institute of Physics and Technology (MIPT)}
\and Alexey Skrynnik\thanks{Artificial Intelligence Research Institute (AIRI)}
\and Shrestha Mohanty\footnotemark[5]
\and Zoya Volovikova\thanks{ITMO University}
\and Julia Kiseleva\thanks{Microsoft Research}
\and Artur Szlam\thanks{Meta AI}
\and Marc-Alexandre C\^ot\'e\footnotemark[5]
\and Aleksandr I. Panov\footnotemark[3]
}
\maketitle

\begin{abstract}
   We present the IGLU Gridworld: a reinforcement learning environment for building and evaluating language conditioned embodied agents in a scalable way. The environment features visual agent embodiment, interactive learning through collaboration, language conditioned RL, and combinatorically hard task (3d blocks building) space. 
\end{abstract}

\section{Introduction}
\label{sec:intro}

In this work, we propose a new reinforcement learning environment called Interactive Grounded Language Understanding (IGLU) Gridworld\footnote{\href{https://github.com/iglu-contest/gridworld}{https://github.com/iglu-contest/gridworld}}. This environment is at the core of the IGLU 2022 competition hosted at NeurIPS\footnote{\href{https://www.iglu-contest.net/}{https://www.iglu-contest.net/}} \cite{IGLU22}.  

\begin{figure}[t]
\centering 
    \includegraphics[ width=\linewidth,clip]{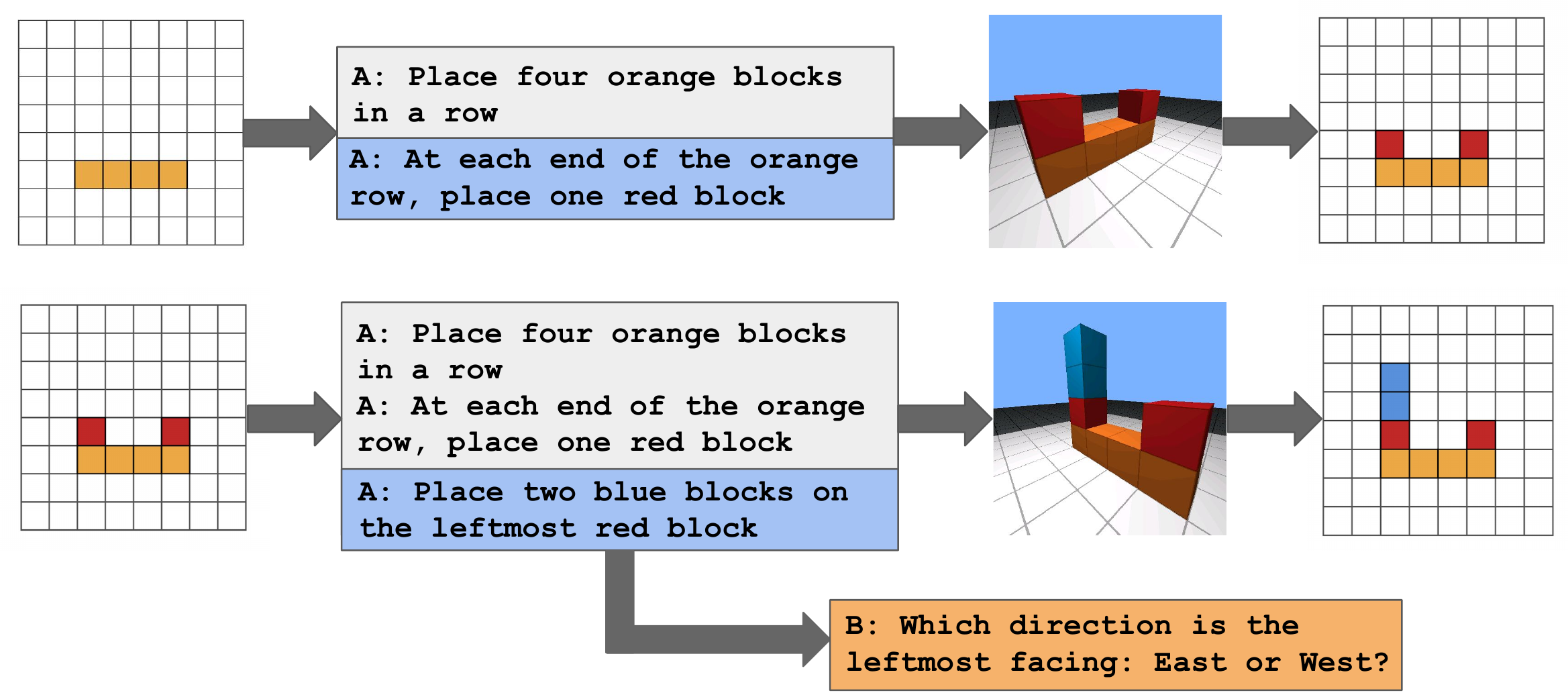}
   \caption{Example of an Architect-Builder task.}
   \vspace{-0.7cm}
   \label{fig:architect-builder}
\end{figure}

The environment consists in an asymmetric collaboration between the \textit{Architect} who has oracle access to the target structure and has to provide an instruction to the \textit{Builder} that has to follow the instruction in a 3D blocks gridworld (Figure~\ref{fig:architect-builder}). While the overall task is \textit{trivially-sided} (i.e., the collaboration starts and ends with simply a small integer 3D array with just 7 possible block colors), it encompasses a nontrivial process which involves contextualized collaborative instructions and an embodied behavior. The IGLU Gridworld enables research in 1) collaborative learning (due to the nature of the task); 2) different kinds of generalization in imitation/reinforcement learning (since the space of tasks is large and grids/dialogues can be easily augmented); 3) hierarchical RL (given the color-shape patterns in structures and the contextualized nature of dialogs); 4. skill learning (since building different sorts of structures requires different action patterns); 5) open-ended learning (since the overall input and target, a 3D grid, is easily formalizable for a co-evolution of the Architect and Builder); and 6) embodied AI (since the environment represents an ego-centric agent that lives in a 3D visual world with combinatorically rich space of states). IGLU Gridworld is implemented in Python to facilitate transparency and simplicity of the game, flexibility of the rules, and uniformity with our data collection tool \cite{Kiseleva2022InteractiveGL}. The last one is especially important for collecting the dataset with behaviours that are valid for our environment.
The contributions of this work are two fold: 1) IGLU Gridworld, a new lightweight and fast RL environment written in pure Python for learning embodied agents that follow dialog instructions with collaborative context; and 2) we propose a multitask hierarchical RL baseline and evaluate its performance on various tasks from \cite{narayan-chen-etal-2019-collaborative} grouped by embodiment skills needed to solve that tasks.
\vspace{-10pt}
\begin{table}[htb]
    \footnotesize
    \resizebox{\columnwidth}{!}{
    \begin{tabular}{
        p{0.11\textwidth}| 
        >{\centering\arraybackslash}p{0.03\textwidth} 
        >{\centering\arraybackslash}p{0.03\textwidth} 
        >{\centering\arraybackslash}p{0.03\textwidth} 
        >{\centering\arraybackslash}p{0.04\textwidth} 
        >{\centering\arraybackslash}p{0.04\textwidth} 
        >{\centering\arraybackslash}p{0.035\textwidth}  
    }
         & \rotatebox{25}{gridworld} & \rotatebox{25}{embodiment} &  \rotatebox{25}{rich task space} & \rotatebox{25}{language (state)} & \rotatebox{25}{API} & \rotatebox{25}{speed (SPS)\footnotemark} \\
        \hline
        Mujoco \cite{todorov2012mujoco} & \xmark & 2D/3D & \xmark & \xmark & C++ & 2.4k  \\
        Baby AI \cite{babyai_iclr19} & 2D & 2D & \cmark & CFG & python & 3k \\
        Nethack \cite{nethack,samvelyan2021minihack,matthews2022skillhack} & 2D & 2D & \cmark & natural & DSL & 14.4k \\
        TextWorld \cite{cote18textworld} & \xmark & \xmark & \cmark & CFG & python & 300 \\
        AI2Thor \cite{ai2thor} & \xmark & 3D & \cmark & natural & python & 30 \\
        Habitat 2.0 \cite{szot2021habitat} & \xmark & 3D & \xmark & \xmark & C++ & 1.4k \\
        Megaverse \cite{petrenko2021megaverse} & 3D & 3D & \xmark & \xmark & C++ & 327k \\
        Xland \cite{xland} & 3D & 3D & \cmark & CFG & \xmark & 30 \\
        MineRL \cite{minerl} & 3D & 3D & \cmark & \xmark & python & 180 \\
        \hline
        IGLU Gridworld & 3D & 3D & \cmark & natural & python & 4.4k
    \end{tabular}
    }
    \caption{IGLU Gridworld vs. related RL environments. CFG: context free grammar. DSL: domain-specific language. Rich task space: can compose tasks to yield a combinatorial task space. API: language that can be used to extend the environment.}
    \label{tab:summary}
\end{table}

\vspace{-12pt}
\section{Related Work}
IGLU Gridworld and IGLU contest \cite{Kiseleva2022InteractiveGL} are inspired by the collaborative task proposed in \cite{narayan-chen-etal-2019-collaborative}. Several environments offer similar features to the ones proposed in IGLU Gridworld. We summarize them in Table \ref{tab:summary}. In particular, IGLU is created for testing 3D embodied agents that are conditioned on natural language instructions. Combined with a rich task space (combination of 3D colored blocks), the environment offers a scalable (i.e., fast and lightweight environment) and extendable (i.e., simple Python API) platform for language and visual embodiment understanding. 

\footnotetext[3]{For Megaverse, we report the one GPU simulation throughput \cite{petrenko2021megaverse}, note that this environment can be only run in parallelization mode whereas, for the rest (including ours), we report single env speed. For XLand, we report the speed from \cite{unity_ai}.}

\vspace{-6pt}
\section{IGLU Gridworld Description}
\vspace{-3pt}
We implement the environment in Python using a simplified version of an open-source Minecraft clone\footnote{\href{https://github.com/fogleman/Minecraft}{https://github.com/fogleman/Minecraft}}. The renderer is decoupled from the core environment and can be disabled with zero overhead enabling the environment to run headless with GPU-acceleration. \textbf{The environment runs at 4400 steps-per-second (SPS) but can reach 17000 when rendering is disabled}. 
Such speed, for a Python RL environment, is possible due to its simplicity.
The observation space consists of
a point-of-view (POV) image $(64, 64, 3)$, an inventory list $(6,)$, a snapshot of the building zone $(11, 9, 11)$, and the agent's position with pitch and yaw angles $(5,)$. The building zone is represented via a 3D tensor with block ids ($0$ for air, $1$ for blue block, etc.). The agent can navigate the building zone, place and break blocks, and switch between block types. The action space is a discrete space of 14 actions: \textit{noop, step [forward|backward|right|left], jump, brake block, place block, choose block type [1-6]} and two-dimensional continuous camera movement action. The reward reflects how complete the built structure is irrespective of its location. We run a convolution-like procedure between the state grid and the target grid. We find the best alignment (translation and rotation of state grid onto the target grid) that yields the maximum per-block intersection\footnote{\href{https://github.com/iglu-contest/gridworld/blob/master/gridworld/task.py}{github.com/iglu-contest/gridworld/blob/master/gridworld/task.py}}. We reward the agent for change in the maximal intersection over time.



\section{Tasks in IGLU Gridworld}

We split tasks into subtasks in IGLU Gridworld using a notion of collaboration \textit{segments}. Each segment is represented by a (collaborative context, target) pair. Context is a (dialog of instructions and clarifying questions, blocks placed in response to that dialog) pair. Target is a (most recent instruction, target blocks placed in response to that instruction) pair. For each task, the environment is initialized with context blocks and the agent is tasked to extend or shrink it to the target blocks. The only source of target information provided to the agent is the target instruction and the context dialog. The segments serve as building blocks for each research direction mentioned in Section \ref{sec:intro}. As one of them, we describe the interactive collaboration used in the IGLU competition 2022. Each task is a sequence of segments and the environment always starts with a segment with empty starting world. For each segment, the agent acts and once the segment is ``solved'', the environment internally switches to the next segment until the last one is solved. We view this process as resembling to the teacher forcing technique used in NLP. In contrast to the evaluation, during training we can reset the env at any segment. 

\section{Baseline}


Training an agent to build any language-defined structure is a challenging task. 
To overcome this, we have developed a multitask hierarchical builder (MHB) with three modules: task generator (NLP part), subtask generator (heuristic part), and subtask solving module (RL part). 

First, we finetuned on the dialog dataset \cite{narayan-chen-etal-2019-collaborative} using BERT-based \cite{bert} 3D transposed convolutional head for the target prediction. 
Second, we developed the subtasks heuristic (with predefined blocks order) module, which uses the target (3D voxel) as input to predict next cube to add or remove.
Third, we trained the agent, using the high-performance APPO algorithm \cite{petrenko2020sf} extended for goal-based policy. The agent curriculum generated compact goal structures (not related to dataset \cite{narayan-chen-etal-2019-collaborative}). And finally, the complete MBH algorithm was evaluated on all tasks from the dataset \cite{narayan-chen-etal-2019-collaborative} (see Table \ref{table:baselines-eval}). We also report the results for the random agent. We labeled each task with the embodied skills required to solve that task. For the skill description and labeling, see gridworld repository \footnote{\href{https://github.com/iglu-contest/gridworld/tree/master/skills}{https://github.com/iglu-contest/gridworld/tree/master/skills}}.


\begin{table}[ht!]
\resizebox{\columnwidth}{!}{
\begin{tabular}{llllll|l}
\toprule
    $F_1$ score & flying	 &tall	&diagonal	&flat	&tricky	&all\\
\midrule
MHB agent (NLP)  & 0.292	 & 0.322	 & 0.242 & 0.334 & 0.295  & 0.313 \\
  MHB agent (full)  & \textbf{0.233}	 & \textbf{0.243}	 & \textbf{0.161} & \textbf{0.290} & \textbf{0.251}  & \textbf{0.258} \\
  Random agent (full) & 0.039 & 0.036	 &0.044	 &0.038	 &0.043	 &0.039  \\

\bottomrule

\end{tabular}
}
\caption{Per-skill aggregation of the baselines performance metrics. For each task, we calculate $F_1$ score between built and target structures. For each skill, we average the performance on all targets requiring that skill. }
\label{table:baselines-eval}
 \end{table}

\vspace{-14pt}
\section{Conclusion} 
\vspace{-3pt}
In this work, we presented IGLU Gridworld, a simulator for testing embodied agent in interactive blocks building task with dialog context. The environment offers its speed and scalability for many research directions related to RL, Embodied AI, NLP, Lifelong learning. We presented an end-to-end multitask hierarchical RL baseline, with notable performance which is higher than the performance of any solution of the IGLU competition at NeurIPS 2021 \cite{Kiseleva2022InteractiveGL}.
 

{\small
\bibliographystyle{ieee_fullname}
\bibliography{PaperForReview}
}

\end{document}